\documentclass[letterpaper, 10 pt, conference]{ieeeconf}

\IEEEoverridecommandlockouts
\usepackage{times}

\usepackage{graphicx}
\usepackage{amsmath}
\usepackage{amsfonts}
\usepackage{color}
\usepackage{url}
\usepackage{amssymb}
\usepackage[noadjust]{cite}
\usepackage{multirow}
\usepackage[linesnumbered,ruled]{algorithm2e}

\usepackage{amsthm}
\usepackage{subfigure}
\usepackage{mathtools}

\usepackage{todonotes}

\DeclareMathOperator*{\argmax}{arg\,max}

\newcommand{\ie}{\textit{i.e.}}
\newcommand{\etal}{\textit{et al}.}
\newcommand{\Continue}{\texttt{continue}}
\newcommand{\Stop}{\texttt{stop}}

\title{\LARGE \bf
Learning When to Quit:  Meta-Reasoning for Motion Planning 
}

\author{
Yoonchang Sung, Leslie Pack Kaelbling, and Tom\'{a}s Lozano-P\'{e}rez
\thanks{We gratefully acknowledge support from NSF grant 1723381; from AFOSR grant FA9550-17-1-0165; from ONR grant N00014-18-1-2847; from the Honda Research Institute, and from MIT-IBM Watson AI Lab.}%
\thanks{The authors are with the Computer Science and Artificial Intelligence Laboratory,
Massachusetts Institute of Technology, Cambridge, MA 02139 USA \texttt{\small \{yooncs8, lpk, tlp\}@csail.mit.edu}.}
}

\begin{document}


\maketitle
\thispagestyle{empty}
\pagestyle{empty}


\begin{abstract}
Anytime motion planners are widely used in robotics. However, the relationship between their solution quality and computation time is not well understood, and thus, determining when to quit planning and start execution is unclear. In this paper, we address the problem of deciding when to stop deliberation under bounded computational capacity, so called \emph{meta-reasoning}, for anytime motion planning. We propose data-driven learning methods, model-based and model-free meta-reasoning, that are applicable to different environment distributions and agnostic to the choice of anytime motion planners. As a part of the framework, we design a convolutional neural network-based optimal solution predictor that predicts the optimal path length from a given 2D workspace image. We empirically evaluate the performance of the proposed methods in simulation in comparison with baselines.
\end{abstract}


\section{Introduction}
\label{sec:intro}
\begin{figure}[ht]
\centering
\includegraphics[width=1.00\columnwidth]{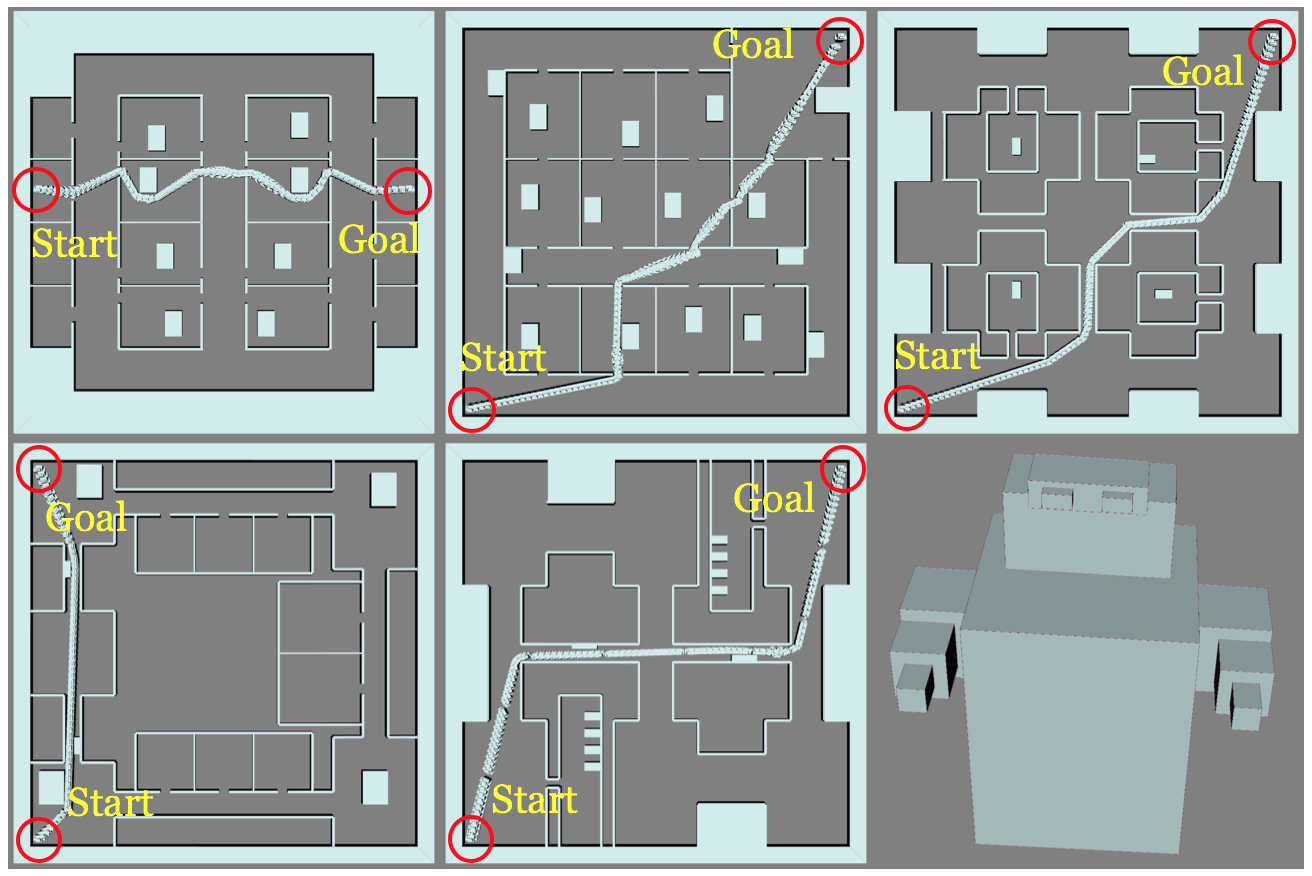}
\caption{Examples of environment distributions and the robot shape used in simulation.
}
\vspace{-1em}
\label{fig:sim}
\end{figure}

Anytime algorithms~\cite{zilberstein1996using} are a class of algorithms that improve solution quality over time and output an answer even if interrupted during computation. Such algorithms occur frequently in time-critical robotic applications and they raise decision-theoretic problems whose objective is to determine when to stop the computation, that is, stop improving the solution, and take action.  In this work, we specifically focus on a class of anytime \emph{motion planning} algorithms~\cite{karaman2011anytime,shome2019anytime,shah2020anytime}. The goal of motion planning (in Figure~\ref{fig:sim}) is to find a collision-free path from the initial configuration to the goal configuration such that the robot does not collide with any obstacles.  Anytime motion planning algorithms typically improve the total length of the path when allowed additional computation time, although one could instead improve another objective, such as obstacle clearance or energy consumption.

Anytime motion planning consist of two phases: (1) finding a feasible path that reaches the goal configuration, and then (2) monotonically improving the path over time. We focus on the second phase in this work:  we consider the problem of determining when to stop deliberation after finding a first feasible path.  For anytime motion planning, we do not typically have a well-characterized relationship between the solution quality and computation time.  Even for asymptotically optimal sampling-based motion planners there does not yet exist a theoretical understanding of their convergence rate to optimality~\cite{gammell4asymptotically}.  Therefore, we will have to rely on data-driven methods to decide when to terminate the computation.

Meta-reasoning~\cite{dean1988analysis,horvitz1990computation,russell1991right} (also known as meta-level control) is reasoning about reasoning, to address decision-theoretic deliberation under bounded computational capacity. Meta-reasoning can be used to determine when to stop deliberation by monitoring the progress of reasoning and regulating the computation time~\cite{ackerman2017meta}. Our objective is to design a meta-reasoning framework for anytime motion planning that applies to any anytime motion planner without modification and generalizes to different environment distributions.  


Our meta-reasoning methods make decisions based on the observed profile of the quality of the current best solution over time (called the \emph{performance profile}).  If this profile is fairly smooth, meta-reasoning can be solved by online prediction without the necessity of data from previous runs; one can extrapolate the future solution quality from a model regressed from the history of solution qualities over the profile at hand~\cite{svegliato2018meta}. This is, however, infeasible in anytime motion planning because performance profiles are generally highly non-smooth due to homotopy-class changes occurring stochastically (see Figure~\ref{fig:example}). Therefore, we will pursue data-driven learning methods to leverage meta-reasoning based data from related planning problems.

We assume the existence of a distribution of problems of interest, which share some similarity in their performance profiles, and assume examples from this distribution are available for training a model that can be used for efficient execution on new instances from that distribution. 
We generate training data by running a given anytime motion planner on the training problems and extracting the resulting performance profiles.
From this data, we learn a decision policy by either model-based or model-free strategies. At execution time, we apply the learned policy to the evolving performance profile of a new problem, assuming that the given problem is drawn from the same distribution as the training problems. Our methods can exploit aspects of the history of solution qualities to determine whether to stop deliberation or keep executing at every decision-making time.

Our model-based approach follows~\cite{hansen2001monitoring}, using a discretization of planning time and solution quality to frame the decision problem as a Markov decision process (MDP).  We learn a transition model for this discrete state and action MDP based on the historical performance profiles in the dataset.  We can then compute an optimal stopping policy for this MDP using value iteration. This simple strategy is effective for small state spaces but does not scale well (in computation and data requirements) as the size of the state space increases, for example, when extending states to include additional features of the history besides the current quality and time. 

We also investigate model-free approaches that use the function-approximation capabilities of neural networks to mitigate the curse of dimensionality. As meta-reasoning is a control problem, approximate dynamic programming or reinforcement learning (RL) can be adopted to learn a policy~\cite{svegliato2020model}. However, we observe that in our problem, we have access to an oracle for the optimal decision policy for each performance profile in the dataset, simply by letting the motion planner run long enough so that we get diminishing returns and determining, {\em post hoc}, the optimal stopping time for each training example. Therefore, model-free meta-reasoning can be formulated as \emph{imitation learning}~\cite{hussein2017imitation}, where the optimal decision policy is treated as an expert demonstration. By exploiting this property, we can take a supervised learning approach, which is generally more robust and has lower sample complexity than RL. We propose two supervised learning-based methods: (1) neural network-based classification with hand-designed features and (2) recurrent neural network (RNN) sequence-to-sequence mapping on the raw performance profile.

In order to generalize over problem instances of different difficulty, the performance profiles need to be normalized so that the states, including quality and time, are comparable across problems.  A critical requirement for this normalization is an estimate of the optimal path length for the current problem.
We have designed a predictor for the optimal solution using a convolutional neural network (CNN) to predict a shortest path length for a given 2D workspace image. We use the predicted optimal solution as a normalizer.  Although this approach could potentially be extended to 3D workspaces using voxel grids or point clouds as input, it currently limits the applicability of our methods to problems with planar workspaces.

Our contributions include:
\begin{itemize}
\item the application of machine-learning methods for general-purpose meta-reasoning to the problem of trading off planning and execution time for anytime motion planning;
\item the formulation, implementation, and comparison of model-based and model-free strategies for this problem; and
\item the use of CNNs to predict optimal solution quality from 2D images.
We develop a solution predictor that predicts an optimal path length from a 2D workspace image.
\end{itemize}
We present experimental results demonstrating the effectiveness of these techniques in comparison to a set of baseline strategies. 

\section{Problem Description}
\label{sec:prob}
Given an anytime motion planning algorithm and a dataset of performance profiles from previous motion planning problems, we want to find  a policy $\pi$ that can be used to decide when computation should be halted on a new problem.  The goal is to receive the maximal expected {\em utility}, defined as a function of path quality and computation time. In this paper, we assume that we have a geometric and kinematic description of the robot and that the robot is operating in a 2D workspace, but that the dimension of the configuration space of the robot is unconstrained.  

The anytime motion planner, after having found an initial path, can be thought of as a function that produces a path quality value as a function of running time; this function is monotonically non-decreasing.  We will normalize this function, as described later, so that the running time $t$ is in the range $[0,1]$ and the quality $q$ is also in the range $[0,1]$. 
We are also given a normalized utility $U(q, t)$ that is a function of the normalized quality $q$ and normalized running time $t$.
We will assume that there are $T+1$ evenly spaced decision steps, $i\in [0,T]$.
At every time step $i$, the performance history $H_i$ is a vector of all the normalized utility values up to that time step.  The robot picks an action $a_i = \pi(H_i)$ at each decision time step $i$ in the current, unseen, workspace. 
If the robot chooses $a_i$ to be \Continue, the deliberation continues to the $(i+1)$-th time step and repeats the same process at time $t_{i+1}$. The meta-reasoning process terminates if $a_i=$ \Stop{} or $i = T$.

We seek a utility-maximizing stopping policy $\pi(H_i)$.  
An ideal policy with oracular information would output $a_i=$ \Stop~for $i = \argmax_{j \in [0,T]}U(q_j, t_j)$.  We can at best seek a policy that is optimal in expectation given the performance profile so far, which is to stop if we do not expect any future time to have a higher utility than the current one:
\begin{multline*}
\pi(H_i) = \text{\Continue\;\; iff} \\
\exists {j \in [i+1,T]},\;\mathbb{E}\big[U(q_j, t_j) \mid H_i\big] > U(q_i, t_i)\;\;.    
\end{multline*}

Let the time at which the first feasible solution is found be, without loss of generality, $0$. The {\em running time} is the maximal allowed time for attempting to improve the solution, at which time meta-reasoning always terminates the algorithm. 
We define a normalized time $t$ as:
\begin{equation}~\label{eqn:time_normalizer}
t=\frac{{\rm current\ time}}{{\rm running\ time}}\in[0, 1].
\end{equation}
We discretize the normalized time range into $T+1$ values.

The solution quality, denoted by $q$, will also be normalized.
We use the optimal (\ie, shortest) path length for this problem and the worst (\ie, longest) path length for this problem for normalization, so that the normalized solution quality $q$ is:
\begin{equation}~\label{eqn:quality_normalizer}
q=\frac{{\rm worst\ path\ length}-{\rm current\ path\ length}}{{\rm worst\ path\ length}-{\rm optimal\ path\ length}}\in[0, 1]\;\;.
\end{equation}
We discretize the range of normalized solution qualities $q$ into $Q+1$ values.

The purpose of the normalization is to make performance histories across different problems comparable and thus enhance our ability to learn a stopping policy that generalizes across problems.
Note that the worst path length is known and corresponds to the length of the first feasible path. The optimal path length, however, is unknown in a previously unseen problem and therefore must be predicted. We describe a CNN-based predictor for the optimal path length in Section~\ref{sec:predictor}. 

The history of normalized solution qualities $H_i$ is the {\em performance profile}, represented by a vector of real values: $(q_0,\ldots,q_i)$.  Note that the solution quality in a performance profile cannot decrease as the deliberation time increases and so $q_i\le q_{i+1}$.

The utility function $U$ encodes the user's trade-off between solution quality and computation time.
The utility function $U(q, t )$ takes as input a normalized solution quality $q$ and normalized time $t$ and outputs a real-valued utility value. Although other functional forms for $U$ are possible, we focus on a linear functional form, \ie, $U(q, t)=w q-(1-w)t$ where $w\in[0,1]$ is a user-defined parameter. Thus, $U(q, t)\in[-1,1]$ due to normalization. This linear functional form has been widely used in the meta-reasoning literature~\cite{russell1991right,hansen2001monitoring}.  Note that, although $q$ always increases, utility may become negative as the computation time becomes large; this is the main motivation for meta-reasoning.

\section{Motivating Example}~\label{sec:motivation}
\begin{figure}[htb]
\centering
\subfigure[Example problem.]{\includegraphics[width=0.36\columnwidth]{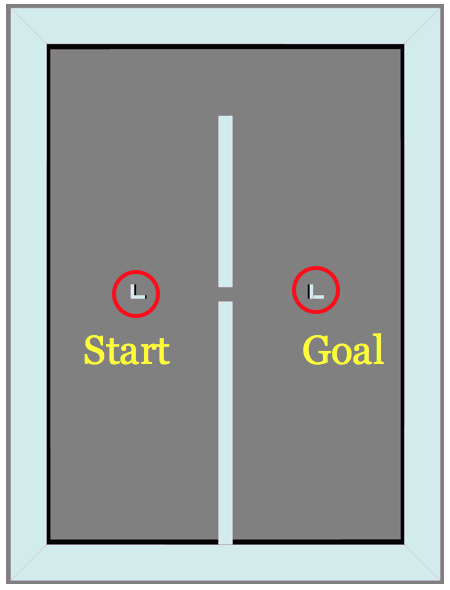}}
\subfigure[Plot of performance profiles.]{\includegraphics[width=0.62\columnwidth]{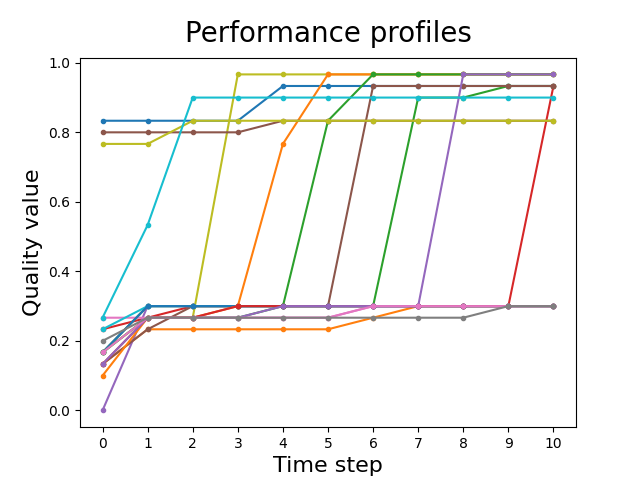}}
\caption{Motivating example where there are two passages in the 2D workspace: the narrow passage in the middle and the wide passage on the top that makes the robot detour around the wall.
A rigid L-shaped robot is used here that can translate in the XY plane and rotate around the Z axis. 
Twenty performance profiles are obtained from twenty randomly generated paths by running the RRT$^*$ algorithm~\cite{karaman2011sampling}. 
}
\vspace{-1em}
\label{fig:example}
\end{figure}

Figure~\ref{fig:example} illustrates a motivating example, in which we are using the anytime motion-planning algorithm RRT$^*$~\cite{karaman2011sampling} in a simple workspace that could occur as part of a larger domain.  Due to the narrow passage located in the middle of the wall (\ie, well-known challenge in sampling-based motion planning~\cite{hsu1998finding}), the algorithm tends to first find a longer path going through the wide passage and then later find a shorter path that passes through the narrow passage.  Therefore, we expect that typical performance profiles would have a single jump when switching from the wide-passage path to the narrow-passage path (\ie, when the homotopy class changes).  However, other performance profiles are also possible because of the stochastic nature of the algorithm. For example, some paths may find the narrow passage first and some may not find the narrow passage within the allotted time. Even the length of the paths that pass through the same passage may also vary and the jump between the same two passages may occur at different times.

 A sampling of 20 executions of the algorithm with the same start and goal configurations illustrates that the performance profiles are highly varying even in this simple example.  Our goal is to design a predictable meta-reasoning framework that learns from such a distribution of performance profiles to find a utility-maximizing stopping policy.

\section{Meta-Reasoning Frameworks}~\label{sec:meta-reasoning}
\begin{figure*}
\centering
\includegraphics[width=\textwidth]{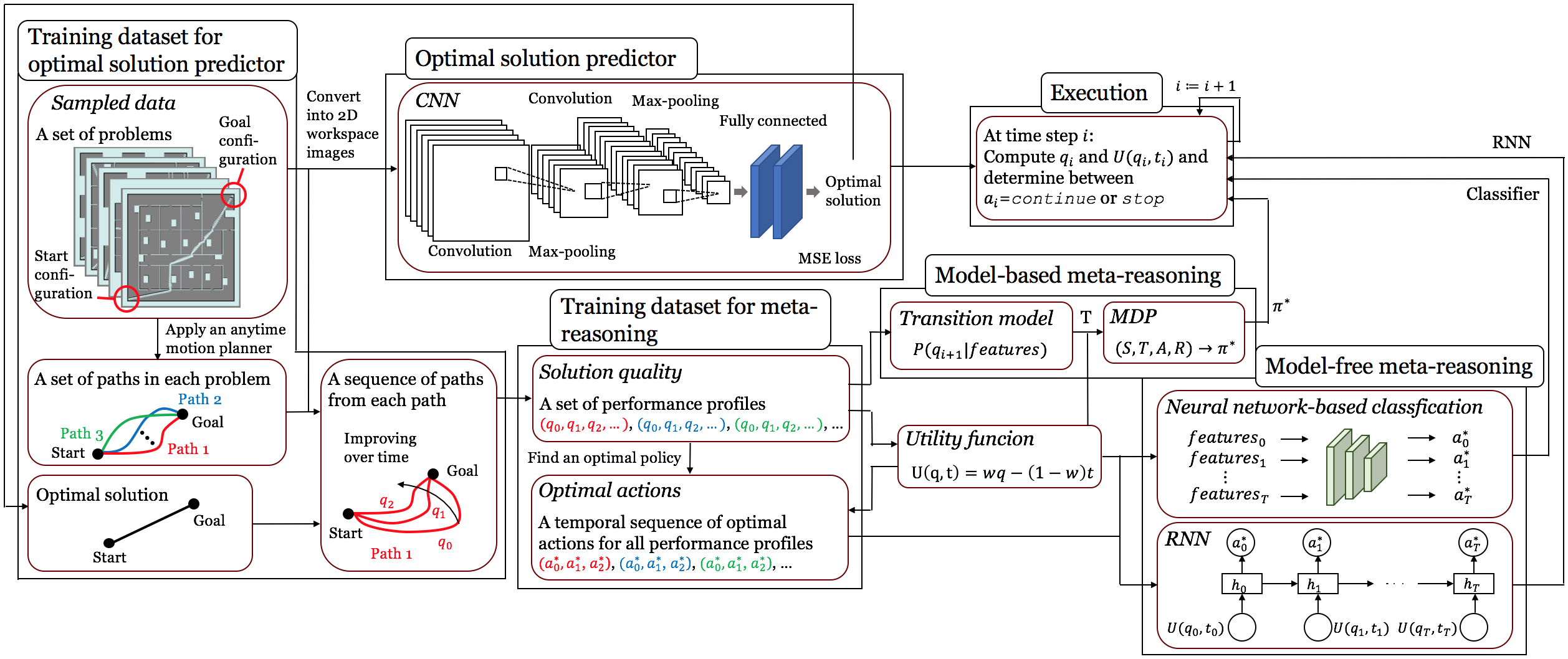}
\vspace{-2em}
\caption{Overall meta-reasoning framework.
}
\vspace{-1em}
\label{fig:framework}
\end{figure*}

In both the model-based and model-free meta-reasoning frameworks, we assume we are given a dataset of representative problems.  For model-based meta-reasoning, during training, we learn an MDP transition model over a state space that represents aspects of the performance profile so far. Then, we can compute a stopping policy via value iteration. This model-based learning and decision process is described in Sections~\ref{subsec:model}.  For model-free meta-reasoning, we directly learn a stopping policy, either in the form of a fully-connected neural network classifier operating on features of the performance profile  or in the form of an RNN on the full performance profile; this process is described in Section~\ref{subsec:free}.  As we mentioned earlier, we also need a predictor during training and testing to estimate an optimal solution quality to be used as a normalizer; the predictor is described in Section~\ref{sec:predictor}. Figure~\ref{fig:framework} shows the overall framework used for meta-reasoning.  We first show how to generate a dataset of performance profiles before we present model-based and model-free meta-reasoning methods.


\subsection{Dataset Generation}~\label{subsec:data}
For each of the training problems, we run the given anytime motion planner for long enough to obtain (with high probability) the optimal quality solution. We use this optimal solution to train the optimal solution predictor (Section~\ref{sec:predictor}). We also run the planner multiple times in each problem to gather a set of performance profiles, each of which consists of a temporal sequence of solution qualities $(q_0, \ldots, q_T)$.
A set of performance profiles forms a training dataset of the model-based approach. For model-free meta-reasoning, however, we additionally need oracular decisions.

For each performance profile and a predetermined utility trade-off $w$ that determines a functional form of the utility function, $wq_i-(1-w)t_i$, we can compute a temporal sequence of optimal actions represented by $(a^*_0,\ldots,a^*_T)$, where 
$a^*_i=$ \Stop~for $i = \argmax_{j \in [0,T]}U(q_j, t_j)$.
Optimal actions before the time step $i$, $(a^*_0,\ldots,a^*_{i-1})$, are \Continue, and those after the time step $i$, $(a^*_{i+1},\ldots,a^*_T)$, are \Stop.
The training dataset of the model-free approach consists of these paired performance profiles and optimal action sequences, and the parameter $w$, with multiple examples constructed from each training problem.

\subsection{Model-Based Meta-Reasoning}~\label{subsec:model}
We can formulate the model-based meta-reasoning problem as an MDP, $(\mathcal{S},\mathcal{A},T,R)$, with: state space $\mathcal{S}$; action space $\mathcal{A}$; transition model $T(s_i,a_i,s_{i+1}) =P(S_{i+1}=s_{i+1}|S_i=s_i, A_i=a_i)$ where $s_i,s_{i+1}\in \mathcal{S}, a_i \in \mathcal{A}$, and $S_i,A_i$ are random variables denoting the state and action taken at time step $i$; and reward function $R(s_i, a_i) =r_i\in \mathbb{R}$.

In our setting, the state at time step $i$ is typically defined to be a tuple of a normalized solution quality and normalized time, $(q_i, t_i)$. 
We will also explore including additional features of the history in the state.
The  set of actions is $\{$\Continue, \Stop$\}$.   The reward function provides a terminal reward based on the utility when the reasoning is stopped, so $R((q_i, t_i), a_i) = wq_i - (1-w)t_i$ if $a_i = \Stop$ and $0$ otherwise.

We first learn the transition model from the dataset of performance profiles. We then compute an optimal policy using this learned transition model.

\noindent {\bf Transition model}\quad
When the state consists of $(q, t)$ pairs, then, because time increments deterministically and sequentially, we only need to explicitly build a transition model of the form $P(q_{i+1} \mid q_i, t_i)$, where we arbitrarily set $q_0$ to be 0. 
%
%

We note that, due to monotonicity in solution quality, $q_{i+1}\geq q_i$, and so we can model the distribution on the next quality value given that the solution quality at time step $i$ is $q_i$, $P(q_{i+1} \mid q_i, t_i)$, 
as having a \emph{categorical} or \emph{multinoulli} distribution with parameter vector $\theta_{t, q}$ in the $(Q+1-Qq_i)$-simplex:  that is, with $Q+1-Qq_i$ positive real parameters that sum to 1, so that $P(q_{i+1} \mid q_i, t_i) = \theta_{t, q}(Qq_{i+1}-Qq_i)$.
Maximum likelihood estimates of the $\theta_{t, q}$ parameters can be computed by counting the number of the corresponding events in the data, and made more robust to sparse data by applying Laplace smoothing.

In general, the $(q, t)$ values are not a sufficient representation of the latent state of the underlying planning process, and prediction accuracy for the next quality value can be improved by conditioning on additional features of the performance profile so far or, indeed, on the whole history 
$(q_0,\ldots,q_{i-1})$. However, increasing the number of conditioning variables
substantially increases the number of parameters in the model and therefore the amount of data required to obtain accurate parameter estimates increases drastically.


Ultimately, the effectiveness of an estimated transition model is borne out in how well the decisions it induces to perform at execution time.  However, it is useful to be able to measure its accuracy as an independent component of the overall system.  We use the {\em negative log likelihood} assigned by our estimated distribution $\widehat{\theta}$ to a set of held-out test data:
\begin{equation}\label{eqn:loss}
{\cal L}(\widehat{\theta}) = \sum_{d \in {\cal D}_\text{test}} -\log(\widehat{\theta}_d)\;\;.
\end{equation}

\noindent {\bf Model-based decision-making}\quad
A policy $\pi$ is a function that maps from state $s$ to action $a$. We compute the policy that is optimal for the estimated transition model, $\pi^*$, by maximizing the expected discounted sum of the utility values over our decision horizon:
$\pi^* =\argmax_{\pi}\mathop{\mathbb{E}}\big[\sum_{i=0}^TR(s_i,a_i)\big]$.
We solve this MDP using value iteration.


\subsection{Model-Free Meta-Reasoning}~\label{subsec:free}
In contrast with model-based meta-reasoning, learning a transition model is not required in the model-free approach. Moreover, the Markovian property assumed in an MDP may not hold for motion planning with non-smooth performance profiles. Thus, we investigate the applicability of the model-free approach that learns a policy directly.  In general, learning a policy requires reinforcement-learning methods, but we are able to take advantage of a special property of this meta-reasoning problem to reduce the policy learning to supervised learning.  

We observe that if, on a training example problem, we run the anytime algorithm and compute the utility of stopping at each time step, we can find the optimal stopping time, $t^*$ and therefore determine, for each time step $i$, that the optimal action $a_i^*$ is \Continue{} if $t_i < t^*$ and \Stop{} otherwise.  These labels constitute a ``demonstration'' of the optimal policy and so allow us to use them as targets for supervised learning.  We explore two supervised learning methods: (1) a feed-forward neural-network classifier and (2) an RNN sequence-to-sequence model. The important distinction is that the RNN model does not require manual feature design. 


\noindent {\bf Neural network-based classification}\quad
From a training dataset consisting of paired performance profiles and optimal action sequences for a predetermined $w$, we design features that can be informative for predicting the optimal decision. Similar to model-based meta-reasoning, besides $q$ and $t$, additional features of the history such as derivatives can be computed. We then learn a classifier to predict the optimal action at time $t_{i+1}$ given features of the performance profile up through time $t_i$ based on a set of labeled examples.  



\noindent {\bf RNN sequence-to-sequence model}\quad
Since our decision whether or not to stop deliberation depends on a sequence of previous observations, it is natural to frame the problem as one of mapping the performance profile so far, as a sequence of utility values to the sequence of actions. We can use an RNN model to learn this sequence-to-sequence mapping, which relieves us from having to hand-craft temporal features, as we did for the feed-forward network.
The training data is a set of input-output pairs, where each pair consists of a temporal sequence of utility values, $U(q_i,t_i)\;\forall i\in[0,T]$, which can be obtained from performance profiles, as input and the corresponding temporal sequence of optimal actions, $a^*_i\;\forall i\in[0,T]$, as output for all performance profiles. 



Because the input and output are low-dimensional, we use a simple RNN with ReLU nonlinearities and cross-entropy loss; we tune the number of hidden dimensions and learning rate using cross-validation.  Although it might be better to use an LSTM, we get very good performance with this very simple model.

\vspace{-0.5em}
\section{Optimal Solution Predictor}~\label{sec:predictor}
One difficulty with this meta-reasoning framework is that performance profiles from problems of different underlying difficulties are pooled together in the training data.  In order to learn anything from that data and apply it to new problems, we would like to normalize any measured solution length in a problem by the {\em actual} shortest path length (length of the optimal solution) for that problem.  Unfortunately, of course, we do not know that optimal path length, so we take the approach of learning to predict it from a description of the problem which, in our case, is a top-down 2D map of the environment.


From a known distribution of problems, we can sample a set of 2D workspace images. For each workspace we run the anytime motion planner for a long term and consider the resulting path length to be the optimum.   The set of paired workspace images and optimal path lengths forms a dataset for a supervised regression problem.
We train a CNN model to learn this mapping. 
We use squared loss and select the learning rate, the number of hidden units and batch size through cross-validation.


\section{Simulation Results}
\label{sec:result}
We perform three evaluation studies, measuring: (1) the effect of the CNN-based solution quality predictor; (2) the accuracy and computational requirements of transition-model estimation; and (3) the value of our approaches for meta-reasoning for anytime motion planning compared to several baselines.  In all simulations, the robot (Figure~\ref{fig:sim}) had three degrees of freedom in motion, that is, 2D translation and rotation. We defined the normalized quality $q$ to be 30 discrete values (\ie, $Q=30$) and normalized running time $t$ to be 200 discrete values (\ie, $T=200$). We set the utility trade-off $w$ to be 0.8 unless stated otherwise. All simulations were implemented based on OMPL~\cite{sucan2012open}.


\subsection{Solution Quality Predictor}~\label{subsec:normalizer}
We analyzed the performance of the CNN-based predictor both in terms of accuracy in predicting optimal path length and in terms of its effect on the overall quality of the meta-reasoning system.
We trained a CNN model composed of two convolutional layers with $(5\times 5)$ kernels interleaved with two $(2\times 2)$ max-pooling layers, followed by fully-connected layers at the end.

\subsubsection{Prediction Accuracy}~\label{subsubsec:accuracy_predict}
\begin{figure}[ht]
\centering
\includegraphics[width=1.00\columnwidth]{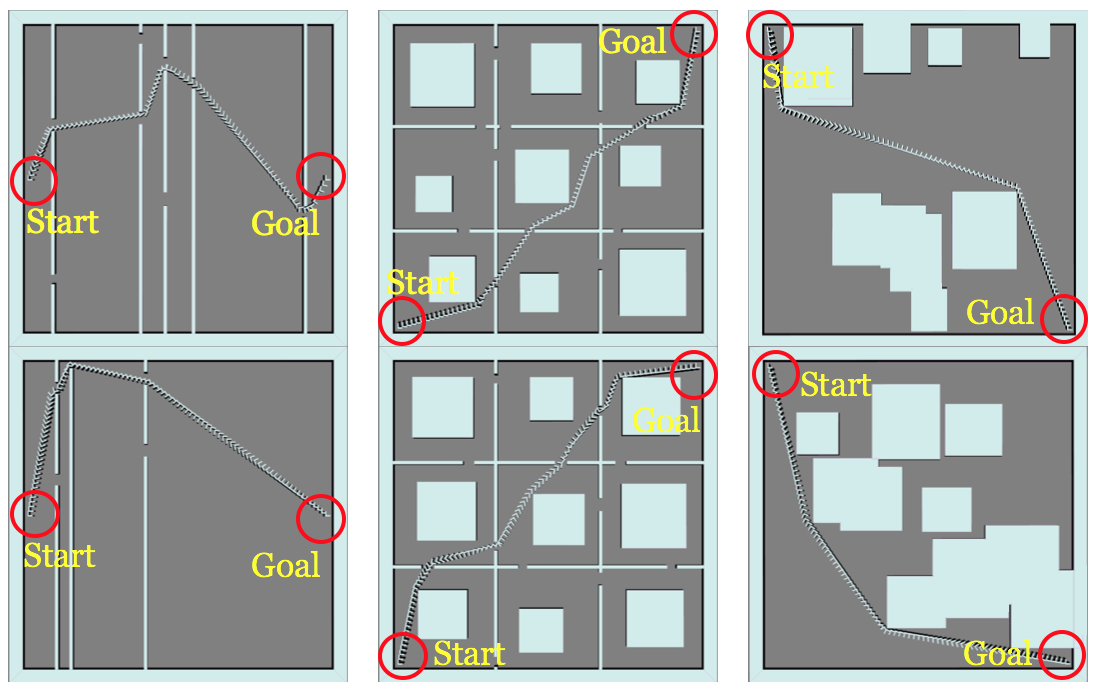}
\caption{The shortest (first row) and longest (second row) paths in three environment distributions (three columns). The lengths of the shortest and longest paths in meters are: 440.49 and 599.32 in the first column distribution, 641.37 and 757.61 in the second column distribution, and 622.91 and 832.13 in the third column distribution.
}
\vspace{-1em}
\label{fig:pred_envs}
\end{figure}

We defined the prediction accuracy to be the percent similarity of the predicted path length to the optimal path length.
To verify the accuracy of the predictor, we constructed an environment distribution by allowing the partial 2D workspace to randomly vary while the start and goal configurations are fixed. We then generated three different distributions having entirely different 2D workspaces. From each environment distribution, we sampled 1000 images (800 images are training data and 200 images are test data) whose resolution is $500\times 500$. 
We ran RRT$^*$ on 800 images for long enough to obtain the shortest path lengths and used them as labels. The accuracy results in Table~\ref{tab:accuracy_predict} were computed after applying a learned CNN model to 200 images that were held out. Figure~\ref{fig:pred_envs} presents the longest and shortest paths from 200 images in each distribution. The results show that larger difference between the longest and shortest path lengths, the lower prediction accuracy. Nevertheless, the CNN-based predictor performed reliably with the amount of variation shown in Figure~\ref{fig:pred_envs}.

\begin{table}[ht]
\centering
\begin{tabular}{|c||c|c|c|}
\hline
Environment type & Env. I & Env. II & Env. III \\
\hline
\hline
Accuracy & 88.21\% & 96.23\% & 94.25\% \\
\hline
\end{tabular}
\caption{Image resolution of environment images and accuracy results of the optimal solution predictor.}
\vspace{-2em}
\label{tab:accuracy_predict}
\end{table}

\subsubsection{Effect of Prediction}~\label{subsubsec:effect_predict}

We validated the importance of having a reasonable predictor in meta-reasoning. To do that, we analyzed the consequence of using different normalizers when finding a stoppinging policy over the same environments above. We considered model-based meta-reasoning and RRT$^*$. We learned a model from 800 workspaces and applied the model to compute the utility values on 200 held-out workspaces. We compared the CNN-based predictor with two other methods: a ground-truth method where the optimal solution was known; and a straight line connecting the start configuration to the goal configuration in 2D workspace which represents a coarse normalizer. 

Table~\ref{tab:normalization} presents the average predicted optimal path length and utility values computed over 200 workspaces for three normalizers. The results imply that poor prediction would decrease the performance of meta-reasoning drastically. The straight-line normalizer produced a comparable performance in the third environment because its prediction was close to the ground truth.

\begin{table*}
\centering
\begin{tabular}{|c||c|c|c||c|c|c|}
\hline
\multirow{2}{*}{Environment type} & \multicolumn{3}{c||}{Average path length} & \multicolumn{3}{c|}{Utility} \\
\cline{2-7} & Ground truth & \ \ \ \ \ CNN\ \ \ \ \  & Straight line & Ground truth & \ \ \ \ \ CNN\ \ \ \ \  & Straight line \\
\hline
\hline
Environment I & 592.42 & \textbf{542.86} & 440.00 & 0.63$\pm$0.09 & \textbf{0.58$\pm$0.08} & 0.39$\pm$0.08 \\
\hline
Environment II & 701.67 & \textbf{716.72} & 622.25 & 0.70$\pm$0.00 & \textbf{0.65$\pm$0.11} & 0.40$\pm$0.00 \\
\hline
Environment III & 694.68 & \textbf{683.66} & 622.25 & 0.67$\pm$0.00 & \textbf{0.66$\pm$0.01} & 0.60$\pm$0.09 \\
\hline
\end{tabular}
\caption{Average optimal path length predicted by three normalizers and utility values obtained by applying the normalizers to meta-reasoning. Utility values are represented by mean and 95\% confidence interval. Three environments are the same as used in Table~\ref{tab:accuracy_predict}.
}
\vspace{-1em}
\label{tab:normalization}
\end{table*}

\subsection{Transition Model Estimation}~\label{subsec:model_result}
We evaluated the accuracy and efficiency of estimating the transition model, using 
4500 trajectories as training data and 500 trajectories as test data obtained from five environment distributions ($500\times 500$ square meters) in Figure~\ref{fig:sim}.

As using all possible features summarizing the entire history is infeasible, we introduce conditioning on two more features besides $q$ and $t$ providing a summary of the performance history:  \emph{slope} and \emph{flatness}.
The slope is just a one-step empirical derivative  $(q_i-q_{i-1})/(t_i-t_{i-1})$ and flatness is the number of previous time steps for which the quality value has been equal to $q_i$.


Table~\ref{tab:accuracy_model} reports the results of estimating the quality transition model conditioned on different feature sets.  The first row reports the data-likelihood loss (\ref{eqn:loss}) on the held-out data.  Adding the slope and flatness features substantially improved the model's predictions.  Unfortunately, as indicated in the second row, which reports the time required to estimate the model parameters, the improvements in accuracy come at a huge computational cost.  This is the curse of dimensionality arising from estimating a large discrete model.



\begin{table*}
\centering
\begin{tabular}{|c||c|c|c|c|c|c|c|}
\hline
Features & No features & $t_i$ & $t_i, q_i$ & $t_i, q_i, q_{i-1}$ & $t_i, q_i, {\rm slope}$ & $t_i, q_i, {\rm slope, flatness}$ & $t_i, {\rm slope, flatness}$ \\
\hline
\hline
Loss & 1.84 & 1.79 & 0.22 & 0.19 & 0.10 & \textbf{0.03} & 1.79 \\
\hline
Time (s) & 0.41 & 2.53 & 10.64 & 2858.63 & 11933.72  & --  & --  \\
\hline
\end{tabular}
\caption{Data-likelihood loss, accuracy, and computation time of estimated transition model for different feature sets. The symbol ``--'' implies the time taken more than a day.}
\vspace{-2em}
\label{tab:accuracy_model}
\end{table*}

\subsection{The Value of Meta-Reasoning for Anytime Motion Planning}~\label{subsec:comparison}
\begin{figure}[ht]
\centering
\includegraphics[width=0.80\columnwidth]{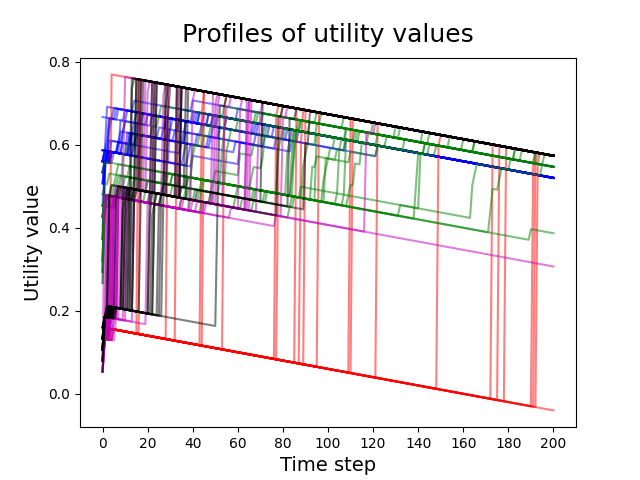}
\caption{Profiles of utility values obtained by running RRT$^*$ over five different environment distributions. Different colors represent different environment distributions.
}
\label{fig:sim_profile}
\end{figure}

We applied the proposed model-based and model-free meta-reasoning frameworks to three anytime motion planners:  RRT$^*$, PRM$^*$~\cite{kavraki1996probabilistic,karaman2011sampling} and Lazy PRM$^*$~\cite{hauser2015lazy}.  We evaluated their effectiveness in five different environment distributions from which we can sample 2D workspaces whose area is $500\times 500$ square meters (Figure~\ref{fig:sim}). We generated a dataset from these distributions and extracted 2D workspace images to learn a CNN-based predictor.  All training was based on  datasets consisting of 4500 training performance profiles and 500 test performance profiles.  We ran the planners 20 times on every held-out workspace to measure the performance of the meta-reasoning methods. In RRT$^*$, we set the range parameter to be large (\ie, 100) so that the planner found a first feasible path quickly but its initial quality was poor. Examples of the profile of the utility values are shown in Figure~\ref{fig:sim_profile}.

\vspace{-1em}

We compared the performance of the proposed methods with other baselines in terms of the mean and 95\% confidence interval of utility values from held-out workspaces while varying the value of $w$. The oracle outputs the optimal actions from all test performance profiles, giving an upper bound on the mean utility value. 

We additionally designed two simple meta-reasoning approaches as baselines: the \emph{fixed-time strategy} and the \emph{fixed-quality strategy}. The fixed-time strategy finds a single time step $i\in[0,T]$ by averaging the time steps at which the optimal \Stop{} action is taken from all training performance profiles, where each optimal time step can be computed by: $\argmax_{j \in [0,T]}U(q_j, t_j)$. Then, in test performance profiles, the fixed-time strategy stops at the chosen time step $i$.
Likewise, the fixed-quality strategy finds an unnormalized quality value $Qq\in[0,Q]$ by averaging the unnormalized quality values at which the optimal \Stop{} action is taken from all training performance profiles. The fixed-quality strategy stops when the performance profile reaches the unnormalized value of $Qq$ in all test performance profiles.

Although we demonstrated that adding features to the model-based method can substantially improve its accuracy, the computational requirements of doing so were impossible for our larger experimental domains. For this reason, 
we used $(q_i, q_{i-1}, t)$ as features for the model-based method. However, $(q_i, q_{i-1}, t, {\rm slope}, {\rm flatness})$ were used for classification. For the classification, we trained a neural network composed of three layers (128, 64, 32 neurons per each layer) and the ReLU activation functions.
For the RNN model, we used 300 neurons in the hidden layer, 0.0001 as a learning rate, and 40000 epochs. 

The results are shown in table~\ref{tab:comparison}.  In general, the model-based meta-reasoning methods provide a substantial advantage over the baselines.  The most reliably good results come from the RNN-based predictor, which is able to discover its own temporal features for prediction and does not suffer from estimation problems due to model size, which cause difficulty for the model-based methods, particularly when increased emphasis is placed on time cost by decreasing $w$. 

\begin{table*}
\centering
\begin{tabular}{|c|c||c|c||c|c||c|c||c|c||c|c||c|c|}
\hline
\multirow{3}{*}{Anytime planner} & Utility & \multicolumn{12}{c|}{Meta-reasoning method} \\
\cline{3-14} & trade-off & \multicolumn{2}{c||}{Oracle} & \multicolumn{2}{c||}{Model-based} & \multicolumn{2}{c||}{Classification} & \multicolumn{2}{c||}{RNN} & \multicolumn{2}{c||}{Fixed time} & \multicolumn{2}{c|}{Fixed quality} \\
\cline{3-14} & (\ie, w) & Mean & CI & Mean & CI & Mean & CI & Mean & CI & Mean & CI & Mean & CI \\
\hline
\hline
\multirow{4}{*}{RRT$^*$} & 0.8 & 0.677 & 0.008 & \textbf{0.663} & \textbf{0.010} & \textbf{0.658} & \textbf{0.014} & \textbf{0.670}$^*$ & \textbf{0.006}$^*$ & 0.575 & 0.016 & 0.634 & 0.011 \\
\cline{2-14} & 0.6 & 0.442 & 0.008 & \textbf{0.414}$^*$ & \textbf{0.012}$^*$ & \textbf{0.399} & \textbf{0.013} & \textbf{0.401} & \textbf{0.013} & 0.341 & 0.014 & 0.337 & 0.016 \\
\cline{2-14} & 0.4 & 0.244 & 0.006 & 0.168 & 0.015 & \textbf{0.195}$^*$ & \textbf{0.008}$^*$ & \textbf{0.191} & \textbf{0.010} & 0.176 & 0.008 & 0.010 & 0.019 \\
\cline{2-14} & 0.2 & 0.096 & 0.004 & 0.031 & 0.016 & \textbf{0.079}$^*$ & \textbf{0.004}$^*$ & \textbf{0.074} & \textbf{0.004} & \textbf{0.072} & \textbf{0.004} & -0.283 & 0.031 \\
\hline
\hline
\multirow{4}{*}{PRM$^*$} & 0.8 & 0.718 & 0.002 & \textbf{0.702}$^*$ & \textbf{0.003}$^*$ & \textbf{0.697} & \textbf{0.003} & \textbf{0.698} & \textbf{0.003} & 0.662 & 0.004 & 0.674 & 0.004 \\
\cline{2-14} & 0.6 & 0.507 & 0.002 & \textbf{0.490}$^*$ & \textbf{0.004}$^*$ & \textbf{0.489} & \textbf{0.003} & \textbf{0.485} & \textbf{0.004} & 0.451 & 0.004 & 0.396 & 0.008 \\
\cline{2-14} & 0.4 & 0.314 & 0.002 & \textbf{0.297}$^*$ & \textbf{0.004}$^*$ & \textbf{0.292} & \textbf{0.003} & \textbf{0.294} & \textbf{0.003} & 0.271 & 0.003 & 0.118 & 0.011 \\
\cline{2-14} & 0.2 & 0.140 & 0.001 & \textbf{0.120} & \textbf{0.004} & \textbf{0.125} & \textbf{0.002} & \textbf{0.125}$^*$ & \textbf{0.002}$^*$ & 0.120 & 0.002 & -0.344 & 0.016 \\
\hline
\hline
\multirow{4}{*}{Lazy PRM$^*$} & 0.8 & 0.643 & 0.006 & \textbf{0.591} & \textbf{0.009} & \textbf{0.593}$^*$ & \textbf{0.008}$^*$ & \textbf{0.589} & \textbf{0.008} & 0.559 & 0.007 & \textbf{0.588} & \textbf{0.009} \\
\cline{2-14} & 0.6 & 0.441 & 0.005 & \textbf{0.376} & \textbf{0.010} & \textbf{0.387} & \textbf{0.006} & \textbf{0.393}$^*$ & \textbf{0.007}$^*$ & 0.361 & 0.006 & 0.302 & 0.010 \\
\cline{2-14} & 0.4 & 0.266 & 0.003 & 0.201 & 0.009 & \textbf{0.224}$^*$ & \textbf{0.004}$^*$ & \textbf{0.221} & \textbf{0.005} & 0.216 & 0.003 & 0.0157 & 0.012 \\
\cline{2-14} & 0.2 & 0.117 & 0.002 & 0.053 & 0.009 & \textbf{0.100} & \textbf{0.002} & \textbf{0.100}$^*$ & \textbf{0.001}$^*$ & \textbf{0.100} & \textbf{0.002} & -0.367 & 0.016 \\
\hline
\end{tabular}
\caption{Comparison with baselines for three different anytime motion planners. In each row, we mark with $*$ the method whose performance is closest to the oracle. CI implies the 95\% confidence interval. Bolded numbers represent the methods whose performance is not statistically significantly different from the method with $*$.}
\vspace{-2em}
\label{tab:comparison}
\end{table*}

\section{Related Work}
\label{sec:related}
The study of meta-reasoning has its roots in the late 80's in the problem of building intelligent decision-making agents under bounded rationality~\cite{russell1991right} in terms of two equivalent models: anytime algorithms~\cite{dean1988analysis} and flexible computation~\cite{horvitz1990computation}.  Here we will focus on meta-reasoning in the context of planning. For more complete surveys, see~\cite{cox2005metacognition,anderson2007review,ackerman2017meta}, and \cite{griffiths2019doing}.

Meta-reasoning has been extensively employed for online planning to decide when to switch to execution during concurrent planning and execution. In online planning, the agent must plan ahead for a bounded time before taking action. The longer the agent spends planning, the higher performance it can achieve but the higher cost it pays, such as wasting energy.
Hay~\etal~\cite{hay2012selecting} applied meta-reasoning as a part of the selection process in Monte-Carlo tree search and compared that with the bandit setting.
Lin~\etal~\cite{lin2015metareasoning} proved the hardness of solving the meta-reasoning version of MDPs and instead proposed an approximation algorithm.
Other planning domains have also adopted meta-reasoning, including algorithm selection in sorting~\cite{lieder2014algorithm} and belief space meta-reasoning in autonomous driving applications~\cite{svegliato2019belief}.

In this work, our interest lies in anytime motion planning in continuous configuration spaces, but there exist previous works on meta-reasoning for path planning in discrete domains.
O'Ceallaigh and Ruml~\cite{o2015metareasoning} designed a meta-reasoning-based search algorithm that dynamically switches between greedy hill-climbing for committing to actions and $A^*$ for deliberation. 
Cserna~\etal~\cite{cserna2017planning} considered durative actions and showed that taking a slower action can lead to a better outcome by allowing a longer time for planning. In their subsequent work~\cite{mitchell2019real}, they presented several backup methods to estimate heuristic values in order to decide which node to expand in tree search.

The most related work to ours is by Hansen and Zilberstein~\cite{hansen2001monitoring}, where they proposed a dynamic-programming approach to solve the model-based variant of meta-reasoning. They later developed online approaches, such as online performance prediction~\cite{svegliato2018meta} and RL-based model-free meta-reasoning~\cite{svegliato2020model} in an attempt to remove the necessity of preprocessing and gathering data. They particularly considered the solution quality to be safety, yielding smooth performance profiles. However, we deal with non-smooth performance profiles due to our definition of path length-based solution quality in this work. As a consequence, a learning framework is necessary for anytime motion planning.


While algorithmic properties like feasibility and completeness have drawn much attention in anytime motion planning work~\cite{gammell4asymptotically}, bounded rationality has generally been overlooked.  However, recent work by Sudhakar~\etal~\cite{sudhakar2020balancing}
formulated an energy-minimization problem that trades off actuation energy and computing energy in anytime motion planners for low-power robotic platforms. Their approach exhibits meta-reasoning in terms of energy consumption although learning from data to help decision-making was not considered. 

\section{Conclusion}
\label{sec:conclusion}
In this work, we introduce a general-purpose meta-reasoning framework for anytime motion planning.  Several data-driven learning methods are proposed, namely model-based and model-free meta-reasoning. We show that learning-based meta-reasoning approaches can improve the utility of anytime motion-planning in embedded settings where computation time must be traded off against execution time.
\vspace{-1em}



\bibliographystyle{IEEEtran}
\bibliography{IEEEabrv,meta-reasoning_refs}


\end{document}